\documentclass{article} 
\usepackage[main, final]{neurips_2026}
\usepackage[utf8]{inputenc}
\usepackage[T1]{fontenc}

\usepackage{amsmath,amsfonts,bm}

\def\eqref#1{equation~\ref{#1}}

\def\1{\bm{1}}

\DeclareMathAlphabet{\mathsfit}{\encodingdefault}{\sfdefault}{m}{sl}
\SetMathAlphabet{\mathsfit}{bold}{\encodingdefault}{\sfdefault}{bx}{n}

\usepackage{hyperref}
\usepackage{url}
\usepackage{graphicx}
\usepackage{subfigure}
\usepackage{booktabs}

\usepackage{amsmath}
\usepackage{amssymb}
\usepackage{mathtools}
\usepackage{amsthm}

\usepackage{multirow}
\usepackage[dvipsnames]{xcolor}
\usepackage{colortbl}
\usepackage[capitalize,noabbrev]{cleveref}
\usepackage{xspace}

\theoremstyle{plain}

\theoremstyle{definition}

\theoremstyle{remark}

\graphicspath{{./}}

\title{Latent-Lens: Visual Perception in Small Language Models Through Latent Communication}

\author{%
  Mo Shahdloo \\
  Microsoft Applied Sciences Group (ASG)
  \And
  Faegheh Sardari \\
  Microsoft Applied Sciences Group (ASG)
  \And
  Mathew Salvaris \\
  Microsoft Applied Sciences Group (ASG)
}

\begin{document}

\maketitle

\begin{abstract}
Bridging independent vision-language models (VLMs) and small language models (SLMs) in a modular multi-agent system typically forces them to exchange information as discrete text, which creates a lossy bottleneck for spatial and fine-grained details. This constraint is especially acute for SLMs, whose limited capacity makes them more reliant on the quality of the input signal. Recent latent-communication methods exchange states between language models. How a VLM can pass visual information to a separately pretrained text-only SLM though remains unexplored. We propose \textbf{Latent-Lens}, a lightweight codec that bridges the representation gap between a frozen VLM and a frozen SLM directly in latent space, enabling them to communicate purely through hidden states. The codec consists of a domain encoder that projects hidden states from the VLM into the SLM's embedding space, and an attention decoder formalised as a LoRA that teaches the SLM to attend to these visual prefixes, while training only ${\sim}1.6$M parameters (0.2\% of the combined model size). On visual question answering (CLEVR, GQA), Latent-Lens achieves $81.0\pm7.9$\% and $65.4\pm5.4$\%, compared with $58.3\pm5.7$\% and $61.8\pm1.0$\% for text communication with the same LoRA budget (+30.2--69.8pp relative to frozen text communication), while on image captioning (Flickr8k) it improves over the same baseline by +16.5 (+18.8 frozen). We also evaluate Latent-Lens with TinyLlama-1.1B and InternVL2-1B, which use different architectures, vocabularies, and hidden dimensions.
\end{abstract}

\section{Introduction}
\label{sec:intro}

Small language models (SLMs) are increasingly deployed on-device and at the edge, where compute and memory budgets can preclude accommodating a monolithic vision-language model (VLM), leaving these systems without visual perception. Tightly coupled VLMs such as LLaVA~\citep{liu2023llava} and Qwen-VL~\citep{bai2023qwenvl} fuse visual and linguistic representations within a single end-to-end trained system. While effective, this tight coupling is inflexible. Upgrading the vision encoder or the language model requires retraining the entire system, and scaling such VLMs down to fit on-device budgets often degrades visual capabilities disproportionately. For example, TinyLLaVA~\citep{zhou2024tinyllava} reports substantial benchmark drops below 1B parameters. Multi-agent approaches offer modularity by pairing a small VLM with a SLM, but have so far relied on text-based communication~\citep{du2023debate}. However, text constitutes a lossy channel. Spatial layouts and fine-grained visual attributes are difficult to verbalise faithfully. SLMs amplify this loss, since with little parametric capacity to fall back on they have no reliable way to recover details the text channel has already discarded. For instance, BeMyEyes~\citep{huang2025bemyeyes} pairs a VLM with a large proprietary LLM through text conversation, but even with substantially more capable models, the generated descriptions inevitably discard fine-grained spatial and visual details.

Recent works on latent multi-agent communication aim to address this bottleneck~\citep{zou2025latentmas,zheng2025thoughtcomm,dery2026kvcache}. However, existing methods operate exclusively within the language modality, and their expressiveness is limited by reliance on representations from a single transformer layer pooling the latent information. Extending these latent-space protocols to cross-modal vision-language multi-agent systems is not yet studied in the literature.

We propose \textbf{Latent-Lens}, a lightweight, model-agnostic codec that bridges the representation gap between a frozen VLM and a frozen SLM, allowing them to communicate directly in latent space and thus bypass the lossy text bottleneck. In contrast to discrete text descriptions (as in BeMyEyes~\citep{huang2025bemyeyes}), our method transmits rich, uncompressed visual representations directly. By extracting multiband latent prefixes, our approach enables cross-modal perception while preserving the depth-wise hierarchy of features encoded across the VLM's layers.

Latent-Lens consists of two functionally distinct components, jointly trained end-to-end. The first is a lightweight MLP domain encoder that extracts and projects hidden states from three semantic bands of the frozen VLM. The second is an attention decoder, instantiated as LoRA adapters on the frozen SLM's attention projections, that teaches the SLM to attend to these projected prefix tokens. We evaluate Latent-Lens with two SLMs, Qwen2.5-0.5B (896-dim~\citep{qwen2024}) and TinyLlama-1.1B (2048-dim~\citep{zhang2024tinyllama}), and two small VLMs, SmolVLM-256M (576-dim~\citep{smolvlm2024}) and InternVL2-1B (896-dim~\citep{chen2024internvl2}), to demonstrate generality across different families of models. Only 1.6-6.1M parameters are trained per pair, representing less than 1\% of the combined frozen parameters in all cases.

We evaluate on free-form VQA tasks using CLEVR~\citep{johnson2017clevr} and GQA~\citep{hudson2019gqa}, and an open-ended captioning task using Flickr8k~\citep{hodosh2013flickr8k}, comparing against VLM-only, SLM-only, text-communication, and LLaMA-Adapter~\citep{zhang2023llamaadapter} deeper-injection baselines. Latent-Lens reaches $81.0\pm7.9$\% on CLEVR, $65.4\pm5.4$\% on GQA, and $18.9\pm1.0$ BLEU-4 on Flickr8k with Qwen2.5-0.5B, and 78.2\% / 66.3\% VQA with TinyLlama-1.1B, with consistent margins over text communication and LLaMA-Adapter across both SLM families.

\section{Related Work}
\label{sec:related}

\paragraph{Text-based inter-model communication.} Multi-agent systems built from independently trained components most commonly exchange information through natural language. \citet{du2023debate} show that multi-round text debate among LLM instances improves factuality and reasoning. \citet{liang2023encouraging} extend this by encouraging divergent thinking across agents to enhance answer diversity, and ChatDev~\citep{qian2024chatdev} applies text-based conversation to coordinate specialised models for software development. In the visual domain, BeMyEyes~\citep{huang2025bemyeyes} pairs a small VLM with a large proprietary LLM, where the VLM describes the image in text and the LLM reasons over the description. While modular, all these approaches are limited by the expressiveness of the text channel, which discards spatial layout and fine-grained visual attributes.

\paragraph{Latent representation transfer between models.} An alternative to text is to communicate directly through hidden states, motivated by the observation that natural language is structurally misaligned with LLMs' high-dimensional representations and causes irreversible information compression~\citep{zhou2025whynl}. In line with this, \citet{yue2025agentprune} show that 28-73\% of text-based inter-agent communication is redundant. The idea originated in multi-agent reinforcement learning. DIAL~\citep{foerster2016dial} introduced differentiable inter-agent messaging, instantiated by CommNet~\citep{sukhbaatar2016commnet} as a mean-pooled hidden-state channel. Subsequent work introduced selectivity over the channel via attention-based targeting in TarMAC~\citep{das2019tarmac} and learnable gating in IC3Net~\citep{singh2019ic3net}, and structured topologies via relational graphs~\citep{jiang2019graph}.

More recently, the principle has been applied to transfer representations between frozen language models. LatentMAS~\citep{zou2025latentmas} shares last-layer hidden embeddings through a latent working memory between \emph{identical} LLMs, achieving up to 14.6\% accuracy gains with lower latency than text debate, and ThoughtComm~\citep{zheng2025thoughtcomm} provides identifiability guarantees for such latent thought exchange. A further line of work targets heterogeneous LLMs with different architectures and vocabularies: Interlat~\citep{du2025interlat} learns compression adapters across architectures for a substantial inference speedup, C2C~\citep{zhang2025c2c} fuses KV caches across Qwen, Llama, and Gemma via learned gating, and \citet{dery2026kvcache} learn KV-cache alignment adapters in the same spirit, SDE~\citep{tang2025sde} takes a hybrid stance, augmenting text with hidden-state deltas, and Vision Wormhole~\citep{liu2026visionwormhole} communicates reasoning traces between VLMs through the receiver's visual-token pathway. None of these methods address transfer from a VLM to a text-only SLM, which is the setting studied here. Moreover, these methods do not use the sender's depth hierarchy as a structured signal. Hidden-state methods exchange a single transformer layer, while KV-cache methods span all layers but pool or align them uniformly rather than separating depth-wise abstraction levels. Finally, although multi-depth structure has been explored within a \emph{single} model - e.g., LLaMA-Adapter~\citep{zhang2023llamaadapter} inserts gated visual residuals at several transformer depths of one VLM - multi-depth transfer across two separate models is unexplored.

\paragraph{Contributions.}
\begin{itemize}
\item We introduce \textbf{Latent-Lens}, a latent-communication codec that passes visual information from a frozen VLM to a separately pretrained text-only SLM.
\item Unlike prior inter-model latent methods that exchange a single transformer layer's representation, Latent-Lens extracts from multiple semantic depth bands of the sender VLM, preserving its perceptual hierarchy.
\item Trained on only 1.6M parameters (0.2\% of the frozen pair), Latent-Lens outperforms text-communication baselines on image-understanding tasks requiring fine-grained visual perception (VQA, captioning), and generalises to alternative SLM and VLM backbone combinations.
\end{itemize}

\section{Methods}
\label{sec:method}

\begin{figure}[t]
\centering
\includegraphics[width=0.9\textwidth]{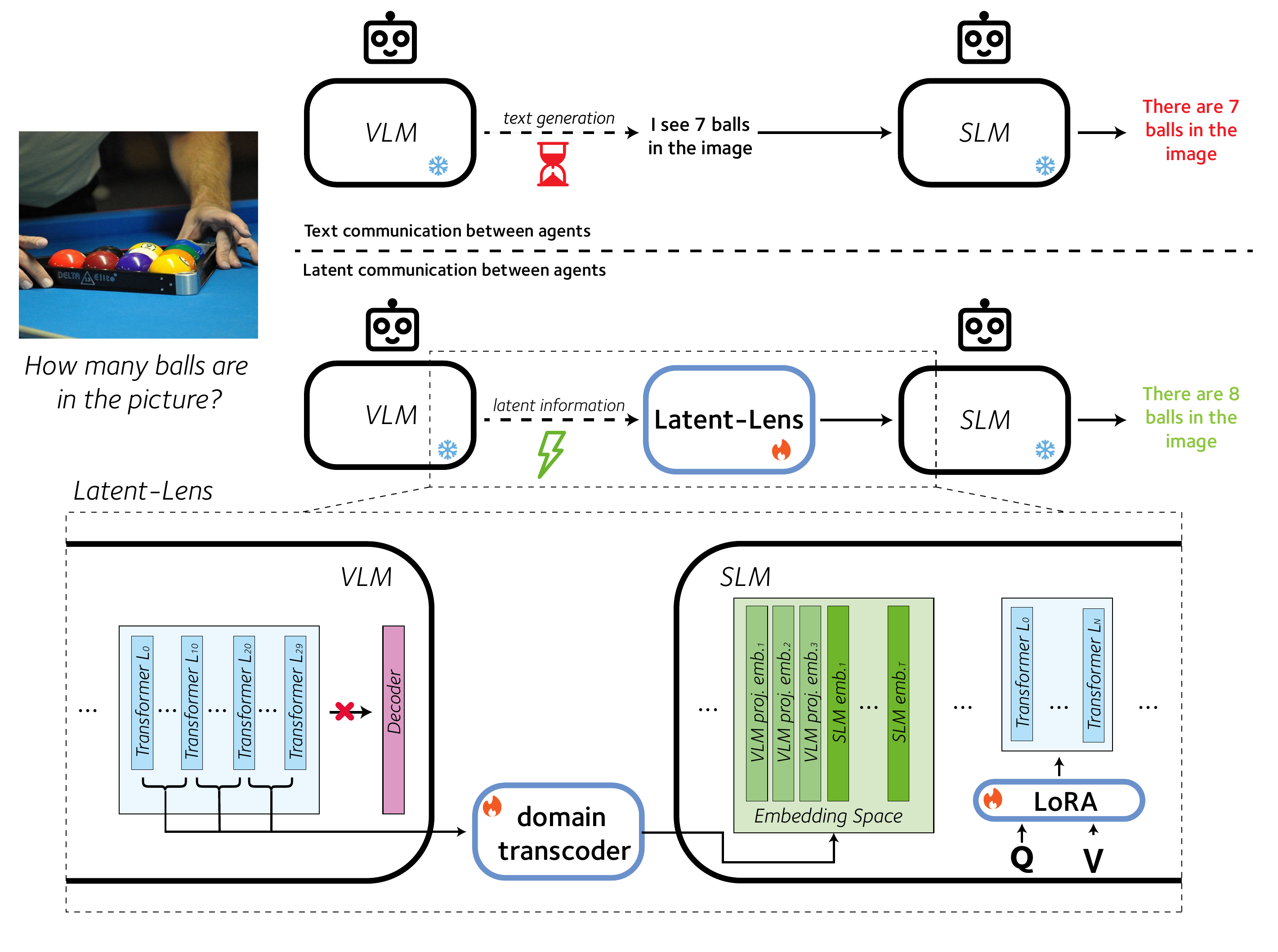}
\caption{\textbf{Latent-Lens architecture.} Text communication passes a generated response from the VLM to the SLM. Latent-Lens passes projected VLM hidden states to the SLM. Black borders denote frozen components. The blue border denotes the trained codec.}
\label{fig:architecture}
\end{figure}

\subsection{Problem Setup}

Given an image and a question, we want a frozen SLM to generate a free-form answer that leverages visual information from a frozen VLM, without any text exchange between the two models (\Cref{fig:architecture}). We denote the VLM by $\mathcal{V}$, with $L_V$ transformer layers and hidden dimension $d_V$, and the SLM by $\mathcal{S}$, with $L_S$ layers and hidden dimension $d_S$. The two have different tokenisers, vocabularies, and embedding dimensions ($d_V \neq d_S$ in general), so the latent channel must explicitly bridge their representation spaces. The specific backbones used in our experiments are described in \Cref{sec:setup}.

\subsection{Multiband Signal Decomposition}

A single-layer representation collapses the VLM's hierarchical processing into one vector, discarding information that earlier or intermediate layers uniquely encode (see \ref{sec:representation_analysis}). To preserve this structure, we propose to decompose the VLM's hidden states across multiple depth bands, each spanning a contiguous group of transformer layers.

To do this, the VLM processes the image and question, producing per-layer hidden states $\{h_\ell\}_{\ell=0}^{L_V-1}$ of the VLM's transformer stack, where $h_\ell\in\mathbb{R}^{T\times d_V}$ is the activation at layer $\ell$ and $T$ is the input token sequence length. Then, we partition the $L_V$ layers into $B\!=\!3$ bands of equal size and compute the band representation as the mean of last-token hidden states within each band:
\begin{equation}
    \mathbf{v}_b = \frac{1}{|\mathcal{L}_b|} \sum_{\ell \in \mathcal{L}_b} h_\ell[:, -1, :] \in \mathbb{R}^{d_V},\quad b \in \{1, 2, 3\}
\end{equation}
where $\mathcal{L}_b$ is the set of layer indices in band $b$. Each band covers a contiguous third of the VLM's transformer stack: band~1 (early), band~2 (mid), and band~3 (late). As we show in \Cref{sec:representation_analysis}, a simple classifier predicts the correct answer more accurately from later bands, while the early and late bands are less similar to each other than the middle and late bands.

\subsection{Domain Transcoding}

The three band representations are projected from the VLM's representation space into the SLM's embedding space using a shared 2-layer MLP that serves as a lightweight domain transcoder.
\begin{equation}
    \tilde{\mathbf{v}}_b = W_2 \, \text{GELU}(W_1 \mathbf{v}_b + b_1) + b_2, \quad W_1 \in \mathbb{R}^{d_S \times d_V}, \; W_2 \in \mathbb{R}^{d_S \times d_S}
\end{equation}
This yields three projected embeddings $\tilde{\mathbf{v}}_1, \tilde{\mathbf{v}}_2, \tilde{\mathbf{v}}_3 \in \mathbb{R}^{d_S}$, one per band.

\subsection{Prefix Delivery and SLM Adaptation}

The transcoded band embeddings are prepended as soft prefix tokens to the SLM's input token sequence:
\begin{equation}
    \mathbf{E}_{\text{input}} = [\tilde{\mathbf{v}}_1; \tilde{\mathbf{v}}_2; \tilde{\mathbf{v}}_3; \mathbf{e}_1, \ldots, \mathbf{e}_T] \in \mathbb{R}^{(3+T) \times d_S}
\end{equation}
where $\mathbf{e}_1, \ldots, \mathbf{e}_T$ are the SLM's text token embeddings. The prefix tokens receive attention mask 1 so the SLM attends to them at every decoding step, but their positions are excluded from the SLM's next-token prediction loss (\textit{i.e.}, cross-entropy loss, label = $-100$).

To adapt the SLM to attend to this novel visual signal, we add LoRA~(rank 4, ablated in \Cref{sec:ablations}) to the query and value projections of every attention layer. The full trainable budget thus consists of the projection MLP plus the LoRA adapters. Per-pair parameter counts are reported in \Cref{app:params} of the Supplementary Materials.

\section{Experiments}
\label{sec:experiments}

\subsection{Setup}
\label{sec:setup}

\paragraph{Models.} The default set of models used here are SmolVLM-256M~\citep{smolvlm2024} ($L_V=30$, $d_V=576$, 256M params) as the VLM agent and Qwen2.5-0.5B~\citep{qwen2024} ($L_S=24$, $d_S=896$, 500M params) as the SLM agent. For this pair, the multiband partition yields layer ranges L0-9 / L10-19 / L20-29 and the trainable budget is 1.6M parameters (1.3M projection MLP + 0.3M LoRA, 0.2\% of the 752M combined). To test generality across heterogeneous backbones (\Cref{sec:generality}), we additionally evaluate Latent-Lens with InternVL2-1B~\citep{chen2024internvl2} ($L_V=24$, $d_V=896$, 900M params) as the VLM agent and TinyLlama-1.1B~\citep{zhang2024tinyllama} ($L_S=22$, $d_S=2048$, 1.1B params) as the SLM agent, spanning two model families (Qwen vs LLaMA), tokenisers, vocabularies, and hidden dimensions. Per-pair specifications and parameter counts are listed in \Cref{app:model_details,app:params}.

\paragraph{Datasets.} We use two free-form VQA benchmarks (\textit{i.e.}, \textbf{CLEVR}~\citep{johnson2017clevr} and \textbf{GQA}~\citep{hudson2019gqa}) and one open-ended captioning benchmark (\textit{i.e.}, \textbf{Flickr8k}~\citep{hodosh2013flickr8k}). \textbf{CLEVR}\footnote{\url{https://huggingface.co/datasets/MMInstruction/Clevr_CoGenT_ValA}}: rendered 3D scenes with questions about counting, spatial relations, and attributes; 5{,}000 examples with a 4{,}000 / 1{,}000 train/test split. \textbf{GQA}\footnote{\url{https://huggingface.co/datasets/lmms-lab/GQA}}: real photographs with compositional questions requiring multi-step reasoning; 9{,}600 / 2{,}400 train/test split. \textbf{Flickr8k}\footnote{\url{https://huggingface.co/datasets/jxie/flickr8k}}: open-ended image captioning; 4{,}000 / 1{,}000 train/test split.

\paragraph{Evaluation.} For VQA we use free-form answer generation and compare predictions to ground truth with a normalized containment heuristic. We lowercase the prediction and reference, remove articles and punctuation, normalize whitespace, and convert number words to digits. We accept an exact match or a match in which every word in the shorter answer appears in the longer answer. Multiple-choice formatting with random distractors inflated SLM-only GQA accuracy in preliminary experiments, motivating the switch to free-form generation as a stricter test of visual understanding. Captioning on Flickr8k is evaluated with BLEU-4 against the reference captions.

\paragraph{Training.} We use AdamW with learning rate $2 \times 10^{-4}$, cosine annealing, weight decay $10^{-4}$, gradient clipping at 1.0. The effective batch size is 16 (4 $\times$ 4 gradient accumulation), and training runs for 5 epochs.

\paragraph{Baselines.} (1)~\textbf{VLM-only (frozen)}: SmolVLM-256M~\citep{smolvlm2024} generates directly from the image and question. (2)~\textbf{SLM-only (frozen)}: Qwen2.5-0.5B~\citep{qwen2024} with no trainable parameters, text input only. (3)~\textbf{Text communication (frozen)}: the VLM produces a textual response to the question, which is prepended to the frozen SLM prompt. (4)~\textbf{SLM-only + LoRA}: same as (2) with 0.3M LoRA~\citep{hu2022lora} parameters (rank 4). (5)~\textbf{Text communication + LoRA}: same as (3) with the same 0.3M rank-4 Q/V LoRA used by Latent-Lens. (6)~\textbf{LLaMA-Adapter}~\citep{zhang2023llamaadapter}: deeper-injection baseline that uses the same shared VLM$\to$SLM projection as our method but additionally injects gated visual residuals at each of the top-12 SLM layers, with a total trainable budget matched to ours (1.6M). (7)~\textbf{Latent-Lens}: our method (1.6M params: 1.3M projection + 0.3M rank-4 Q/V LoRA). The +LoRA variants isolate the effect of the communication channel from task adaptation by giving the SLM the same LoRA budget as in our method, while the LLaMA-Adapter baseline contrasts our input-prefix + LoRA design against layer-wise gated-residual injection at matched parameter budget.

\subsection{Main Results}

\begin{table}[t]
\centering
\caption{Main results on CLEVR, GQA (accuracy) and Flickr8k (BLEU-4). Source column lists the paper introducing each baseline. Trained conditions report mean $\pm$ sample standard deviation over three runs; frozen conditions are deterministic point estimates.}
\label{tab:main}
\small
\setlength{\tabcolsep}{3pt}
\resizebox{\textwidth}{!}{%
\begin{tabular}{llcccc}
\toprule
\textbf{Condition} & \textbf{Source} & \textbf{Params} & \textbf{CLEVR} & \textbf{GQA} & \textbf{Flickr8k} \\
 & & & (\%) & (\%) & (BLEU-4) \\
\midrule
VLM only (frozen) & SmolVLM-256M~\citep{smolvlm2024} & 0 & 58.6 & 41.8 & 13.9 \\
SLM only (frozen) & Qwen2.5-0.5B~\citep{qwen2024} & 0 & 1.5 & 18.0 & 11.8 \\
Text communication (frozen) & this work & 0 & 11.2 & 35.2 & 0.1 \\
\midrule
SLM only + LoRA & \citet{hu2022lora} & 0.3M & $42.4\pm0.2$ & $46.2\pm1.3$ & $12.7\pm0.9$ \\
Text communication + LoRA & this work & 0.3M & $58.3\pm5.7$ & $61.8\pm1.0$ & $2.4\pm2.0$ \\
LLaMA-Adapter & \citet{zhang2023llamaadapter} & 1.6M & $73.6\pm3.5$ & $61.9\pm6.1$ & $18.0\pm0.7$ \\
\midrule
Latent-Lens (ours, $B\!=\!3$) & this work & 1.6M & $\mathbf{81.0\pm7.9}$ & $\mathbf{65.4\pm5.4}$ & $\mathbf{18.9\pm1.0}$ \\
\bottomrule
\end{tabular}%
}
\end{table}

\Cref{tab:main} reports the main comparison on CLEVR, GQA, and Flickr8k. Results show that Latent-Lens outperforms all baselines, with $81.0\pm7.9$\% on CLEVR, $65.4\pm5.4$\% on GQA, and $18.9\pm1.0$ BLEU-4 on Flickr8k.

\paragraph{Frozen baselines.} We first establish a floor by evaluating each component in isolation. The frozen SLM reaches 1.5\%, 18.0\%, and 11.8 BLEU-4 on CLEVR, GQA, and Flickr8k, respectively. The VLM in isolation reaches 58.6\%, 41.8\%, and 13.9 BLEU-4 on CLEVR, GQA, and Flickr8k, respectively. Prepending VLM-generated text responses to the frozen SLM (text communication, frozen) yields 11.2\%, 35.2\%, and 0.1 BLEU-4 on CLEVR, GQA, and Flickr8k, respectively. Latent-Lens reaches $81.0\pm7.9$\%, $65.4\pm5.4$\%, and $18.9\pm1.0$ BLEU-4 on CLEVR, GQA, and Flickr8k, respectively, exceeding frozen text communication by +69.8pp, +30.2pp, and +18.8~BLEU-4. The two VQA scores under text communication remain well below VLM-only direct generation, indicating that the text channel does not preserve the visual information available to the VLM internally. 

\paragraph{Controlling for adaptation.} To disentangle the contribution of the communication channel from that of LoRA task adaptation, we add a 0.3M-parameter LoRA adapter (rank 4, matched to the Latent-Lens budget) on top of both the SLM-only and text-communication baselines. SLM+LoRA reaches $42.4\pm0.2$\% / $46.2\pm1.3$\% / $12.7\pm0.9$ BLEU-4, text-communication+LoRA reaches $58.3\pm5.7$\% / $61.8\pm1.0$\% / $2.4\pm2.0$ BLEU-4, and Latent-Lens reaches $81.0\pm7.9$\% / $65.4\pm5.4$\% / $18.9\pm1.0$ BLEU-4. The residual margin is therefore not attributable to the LoRA capacity but to the content of the channel. Changing only the VLM sender prompt to request one or two concise sentences of question-relevant visual evidence gives 69.5\% on CLEVR and 58.2\% on GQA with text communication + LoRA, still far below Latent-Lens.

\paragraph{Alternative VLM-SLM couplings.} We compare Latent-Lens against LLaMA-Adapter~\citep{zhang2023llamaadapter}, an alternative method for conditioning a frozen language model on projected visual features through gated prompts attended at the upper transformer layers. In our adaptation we retain the shared VLM$\to$SLM projection and replace only the input-prefix + LoRA Q/V adaptation with the LLaMA-Adapter mechanism at the top-12 SLM layers, under a matched 1.6M trainable parameter budget. The baseline reaches $73.6\pm3.5$\%, $61.9\pm6.1$\%, and $18.0\pm0.7$~BLEU-4 on CLEVR, GQA, and Flickr8k, respectively, against $81.0\pm7.9$\%, $65.4\pm5.4$\%, and $18.9\pm1.0$ for Latent-Lens. We attribute the gap to where the visual signal enters the SLM. LLaMA-Adapter introduces visual context only at the upper transformer layers, so the lower-layer text representations are computed without it. Latent-Lens injects the visual prefix at the input embedding, and all subsequent layers compute over visually-conditioned text representations through the SLM's pre-trained self-attention.

\paragraph{Per-benchmark analysis.}
\textbf{CLEVR} isolates spatial relations, counting, and attribute recognition over synthetic 3D scenes, where the latent channel was expected to separate most clearly from text since fine-grained structured visual content is hardest to verbalise. Latent-Lens achieves $81.0\pm7.9$\%, compared with $58.3\pm5.7$\% for text communication + LoRA.
\textbf{GQA} contains compositional questions over real photographs, on which language priors and short descriptions are expected to carry more signal than on CLEVR. Latent-Lens achieves $65.4\pm5.4$\%, compared with $61.8\pm1.0$\% for text communication + LoRA, with SLM+LoRA ($46.2\pm1.3$\%) only modestly above VLM-only (41.8\%), consistent with the partial sufficiency of text-based reasoning at this benchmark.
\textbf{Flickr8k} is an open-ended captioning task that allows the VLM to emit substantially richer text than the terse VQA answers, which we hypothesised would narrow the gap between Latent-Lens and VLM-only generation. The VLM produces 40.6-word descriptions on average against 1.0 and 1.1 words on CLEVR and GQA, and the +5.0 BLEU-4 margin of Latent-Lens over VLM-only is the smallest in our experiments, consistent with this hypothesis.
To examine whether the observed captioning improvement extends beyond BLEU-4, we rescore the saved captions from Latent-Lens and text communication + LoRA on a diagnostic 200-example Flickr8k subset using CIDEr~\citep{vedantam2015cider}, obtaining 0.0587 and 0.0215, respectively.

\subsection{Generality across model backbones}
\label{sec:generality}

To test generality across heterogeneous backbones, we evaluate Latent-Lens on all four pairings of two VLMs (SmolVLM-256M, InternVL2-1B) and two SLMs (Qwen2.5-0.5B, TinyLlama-1.1B). The two VLMs differ in scale and feature dimensionality, while the two SLMs differ in family (Qwen vs LLaMA), tokeniser, vocabulary, and hidden dimension (896 vs 2048). Each pairing only retrains the projection MLP for the relevant VLM-SLM dimensions and the SLM LoRA adapters, leaving both backbones frozen.

Across all four pairings, Latent-Lens exceeds both the VLM-only and SLM+LoRA baselines on the average score (\Cref{tab:generality}). The gain over SLM+LoRA stays in the +15.8 to +26.9 range across all four pairings, indicating that the latent channel adds visual content the text-only SLM cannot recover from its language prior. The smallest gain over VLM-only (+5.5, InternVL2-1B + Qwen2.5-0.5B) occurs when the VLM is already strong in isolation. The Flickr8k scores additionally reflect the SLM's text-only fit to the caption distribution, since TinyLlama-1.1B reaches 27.1 BLEU-4 with no visual input against 11.7 for Qwen2.5-0.5B, which is why the TinyLlama pairings reach higher Flickr8k scores. The InternVL2-1B/TinyLlama-1.1B pairing achieves 87.4\% on CLEVR and 72.4\% on GQA.

\begin{table}[t]
\centering
\caption{Generality across all four VLM $\times$ SLM pairings. Score is the mean of CLEVR (\%), GQA (\%), and Flickr8k (BLEU-4). VLM-only and SLM+LoRA columns are matched-VLM and matched-SLM baselines averaged the same way. These are single-run compatibility results for each pairing. \Cref{tab:main} reports three-run means for the default SmolVLM/Qwen pair.}
\label{tab:generality}
\begin{tabular}{llccc}
\toprule
\textbf{VLM} & \textbf{SLM} & \textbf{VLM-only} & \textbf{SLM+LoRA} & \textbf{Latent-Lens} \\
\midrule
SmolVLM-256M & Qwen2.5-0.5B   & 38.1 & 33.2 & 60.1 \\
SmolVLM-256M & TinyLlama-1.1B & 38.1 & 38.2 & 54.0 \\
InternVL2-1B & Qwen2.5-0.5B   & 51.5 & 33.2 & 57.0 \\
InternVL2-1B & TinyLlama-1.1B & 51.5 & 38.2 & 63.4 \\
\bottomrule
\end{tabular}
\end{table}

\subsection{Representation Analysis}
\label{sec:representation_analysis}
We measure two properties of the multiband latent channel, the task-relevant information contained in each individual band and the overlap between bands.

\paragraph{Per-band information content.}
To estimate how much task-relevant information each depth band carries, we fit a logistic regression probe on each band's frozen representation, predicting the correct VQA answer from the set observed in the training split. Linear-probe accuracy measures the decodability of task-relevant information from each band.

\begin{table}[t]
\centering
\caption{Linear probe accuracy per band on CLEVR and GQA.}
\label{tab:probes}
\begin{tabular}{lcc}
\toprule
\textbf{Representation} & \textbf{CLEVR (\%)} & \textbf{GQA (\%)} \\
\midrule
Band 1 (early, L0-9) & 59.6 & 43.1 \\
Band 2 (mid, L10-19) & 67.9 & 53.2 \\
Band 3 (late, L20-29) & 70.1 & 57.8 \\
\midrule
Last layer only & 68.6 & 56.5 \\
\bottomrule
\end{tabular}
\end{table}

Higher bands carry more task-relevant information (\Cref{tab:probes}), consistent with the depth-abstraction hierarchy~\citep{tenney2019bert}. Prior inter-model latent methods extract from a single transformer layer (typically the last), corresponding to the last-layer probe at 68.6 / 56.5. The Band~3 probe (mean of L20-29) reaches 70.1 / 57.8, above the last layer alone, while the full Latent-Lens system reaches 89.6 / 71.6, indicating that the nonlinear transcoder and LoRA adapter recover cross-band information that a linear readout does not.

\paragraph{Inter-band similarity.}
To assess similarity between the bands, we measure linear CKA~\citep{kornblith2019cka}. CKA between Bands~1 and~3 is 0.2 on GQA and 0.4 on CLEVR, while CKA between Bands~2 and~3 is 0.8 and 0.9.

\paragraph{Effective dimensionality (Latent vs.\ Text).}
To compare the effective dimensionality of the latent and text channels directly, we quantify each via \emph{effective rank}~\citep{roy2007effective}, $r_{\mathrm{eff}} = \exp\bigl(-\sum_i \hat{\sigma}_i \log \hat{\sigma}_i\bigr)$ with $\hat{\sigma}_i = \sigma_i / \sum_j \sigma_j$ the normalised singular values of the centered representation matrix. Per band, $r_{\mathrm{eff}}$ ranges from 187 to 230 on CLEVR and 267 to 364 on GQA out of 576 dimensions. Concatenating the three bands yields 304 on CLEVR and 473 on GQA. By comparison, SmolVLM-256M emits 2.0 tokens per CLEVR question and 2.3 per GQA question on average, e.g.\ ``8.'', ``Yes'', ``Giraffe'' (see \Cref{app:examples}). The latent channel thus has roughly two orders of magnitude more effective dimensions per sample than the text channel.

\subsection{Latency and Efficiency}

\begin{table}[t]
\centering
\caption{Inference latency and GPU memory on a single 24GB GPU (fp16, 20-token generation, averaged over 10 runs).}
\label{tab:latency}
\begin{tabular}{lccc}
\toprule
\textbf{Condition} & \textbf{Total Params} & \textbf{GPU Memory} & \textbf{Latency} \\
\midrule
VLM only & 256M & 1{,}008 MB & 277 $\pm$ 1 ms \\
SLM only (+ LoRA) & 495M & 1{,}013 MB & 620 $\pm$ 36 ms \\
Text communication & 752M & 2{,}113 MB & 2{,}137 $\pm$ 9 ms \\
Latent-Lens & 752M & 2{,}477 MB & 781 $\pm$ 11 ms \\
\bottomrule
\end{tabular}
\end{table}

Replacing autoregressive VLM text generation with a single VLM forward pass should reduce end-to-end inference cost. To quantify this, we measure inference latency and GPU memory under matched generation budgets (\Cref{tab:latency}). Text communication requires the VLM to generate a 50-token description (1{,}523ms) before the SLM begins answering, whereas Latent-Lens extracts the VLM hidden states in a single forward pass (134ms). End-to-end latency is 781ms for Latent-Lens against 2{,}137ms for text communication, a 2.7$\times$ reduction.

\subsection{Ablations}
\label{sec:ablations}

\paragraph{Number of bands.} To assess sensitivity to the depth-band partition, we ablate $B \in \{1, 2, 3, 5, 7\}$ with $r\!=\!4$ fixed. $B\!=\!1$ is the highest-scoring setting on CLEVR (91.8\%) and Flickr8k (19.5 BLEU-4), while $B\!=\!5$ and $B\!=\!7$ are highest on GQA (76.2\% and 76.5\%). The default $B\!=\!3$ yields 89.6\% / 71.6\% / 19.0, within 2.2pp / 4.9pp of the per-benchmark maximum on the two VQA tasks and 0.5 BLEU-4 of the maximum on Flickr8k. $B\!=\!1$ averages the last-token states from all VLM layers into one prefix token. We use $B\!=\!3$ by default, although the highest-scoring value of $B$ differs across tasks.

\begin{table}[t]
\centering
\caption{Comparison of three depth bands with three copies of the final-layer representation. Values are mean $\pm$ sample standard deviation over three runs.}
\label{tab:depth_control}
\small
\begin{tabular}{lcc}
\toprule
\textbf{Representation} & \textbf{CLEVR (\%)} & \textbf{GQA (\%)} \\
\midrule
Final layer $\times3$ & $71.5\pm0.9$ & $62.7\pm1.6$ \\
Three depth bands & $81.0\pm7.9$ & $65.4\pm5.4$ \\
\bottomrule
\end{tabular}
\end{table}

\paragraph{LoRA rank.} To assess whether additional LoRA capacity helps, we ablate $r \in \{4, 8, 16, 32, 64\}$ with $B\!=\!3$ fixed. Rank $r\!=\!4$ yields the highest scores on all three benchmarks (89.6 / 71.6 / 19.0), against 88.0 / 57.1 / 14.9 at $r\!=\!16$ and 86.5 / 70.0 / 14.4 at $r\!=\!64$. Intermediate ranks ($r\!=\!8, 32$) are non-monotonic, with $r\!=\!8$ dropping to 73.0\% on CLEVR. Increasing LoRA capacity beyond rank 4 does not improve any of the three metrics universally.

\section{Discussion}
\label{sec:discussion}

We introduced \textbf{Latent-Lens}, a lightweight cross-modal codec that adds visual perception to a frozen text-only small language model. Where prior multi-agent VLM-SLM systems use natural language as the inter-model channel, Latent-Lens routes the VLM's hidden states directly into the SLM and transmits visual content as a continuous representation rather than a short token description. Discrete text irrecoverably discards spatial layout and fine-grained perceptual attributes that the VLM has already encoded internally, while the latent channel preserves them. The increased expressiveness translates into higher downstream accuracy than text communication at the same parameter budget. Replacing autoregressive image-to-text generation with a single VLM forward pass also keeps the latency overhead over the SLM-only baseline small, so the SLM gains visual perception without paying the cost of describing the image in words. Note however that these opaque messages are harder to inspect and audit compared to text communication. The codec leaves both backbones untouched and transfers across heterogeneous VLM and SLM families with different vocabularies and hidden dimensions. Finally, our experiments are limited to communication between two agents and do not evaluate larger multi-agent systems and the band-count ablation shows that no single value of $B$ performs best across all tasks.

Several extensions are natural follow-ups. The current design uses uniform band partitions over the VLM's depth, so learned or adaptive band boundaries (e.g., MoE-style differentiable routing over depth) are a promising direction. Communication is unidirectional in this work, while bidirectional or multi-turn latent dialogue, in which the SLM can request clarification and the VLM can revise its prefix accordingly, is a natural next step. Broader evaluation across other vision-language tasks (visual grounding, fine-grained recognition) and integration with KV-cache alignment on the language side~\citep{dery2026kvcache,zhang2025c2c} would further stress-test and extend the approach.

\section*{Funding and Competing Interests}

Microsoft supported this work through the authors' employment. The authors declare no competing interests beyond their employment at Microsoft.

\bibliographystyle{plainnat}
\bibliography{references}

\clearpage
\appendix

\section*{\LARGE Supplementary Material}
\label{sec:appendix}

\section{Model Details}
\label{app:model_details}

\begin{table}[h]
\centering
\caption{Model specifications and band assignments.}
\begin{tabular}{lcccc}
\toprule
\textbf{Model} & \textbf{Layers} & \textbf{Hidden dim} & \textbf{Params} & \textbf{Band ranges ($B\!=\!3$)} \\
\midrule
SmolVLM-256M (VLM) & 30 & 576 & 256M & L0-9 / L10-19 / L20-29 \\
InternVL2-1B (VLM) & 24 & 896 & 900M & L0-7 / L8-15 / L16-23 \\
Qwen2.5-0.5B (SLM) & 24 & 896 & 500M & - \\
TinyLlama-1.1B (SLM) & 22 & 2048 & 1.1B & - \\
\bottomrule
\end{tabular}
\end{table}

\section{Trainable Parameter Breakdown}
\label{app:params}

\begin{table}[h]
\centering
\caption{Trainable parameters of Latent-Lens for the Qwen2.5-0.5B pair.}
\begin{tabular}{lrl}
\toprule
\textbf{Component} & \textbf{Params} & \textbf{Description} \\
\midrule
Domain encoder & 1{,}300K & 2-layer MLP: $576 \to 896$ (GELU) $\to 896$ \\
Attention decoder (LoRA) & 300K & Rank 4, q/v projections, all 24 layers \\
\midrule
\textbf{Total} & \textbf{1{,}600K} & 0.2\% of frozen parameters \\
\bottomrule
\end{tabular}
\end{table}

\section{Evaluation Details}
\label{app:eval}

Free-form answers are compared via a normalised string match. The prediction and the reference are lowercased, stripped of leading articles (a, an, the) and punctuation, and number words are mapped to digits (e.g., ``three'' $\to$ ``3''). A prediction is accepted on either an exact match against the reference or bidirectional word-set containment, where every word in the shorter answer appears in the longer one. The metric was validated against 50 hand-labelled CLEVR and GQA examples with 100\% agreement.

Early experiments used multiple-choice formatting with random distractors, which inflated SLM-only GQA accuracy from 56.5\% to 96\% because the model could eliminate implausible distractors without any visual understanding. The switch to free-form generation was essential for measuring actual visual comprehension.

\section{Text-Communication Prompts}
\label{app:text_prompts}

The following templates specify the task text supplied to each agent. For VQA, \texttt{\{question\}} is the dataset question; for image captioning it is \texttt{Describe this image.} The VLM receives the image together with its prompt through the backbone's chat template. In text communication, \texttt{\{vlm\_description\}} is the VLM's generated response.

\paragraph{Default VLM sender: VQA.}
\begin{quote}\small
\texttt{Question: \{question\}}\\
\texttt{Answer with a single word or short phrase:}
\end{quote}

\paragraph{Default VLM sender: Image Captioning.}
\begin{quote}\small
\texttt{Question: Describe this image.}\\
\texttt{Provide a detailed caption:}
\end{quote}

\paragraph{Text-communication SLM receiver: VQA and Image Captioning.}
The same short-answer instruction is used for both tasks.
\begin{quote}\small
\texttt{A vision model looked at the image and responded: "\{vlm\_description\}"}\\
\texttt{Based on this, answer the following question with a single word or short phrase.}\\
\texttt{\{question\}}\\
\texttt{Answer:}
\end{quote}

\paragraph{SLM-only and Latent-Lens receiver: VQA.}
\begin{quote}\small
\texttt{Answer the following question with a single word or short phrase.}\\
\texttt{\{question\}}\\
\texttt{Answer:}
\end{quote}

\paragraph{SLM-only and Latent-Lens receiver: Image Captioning.}
\begin{quote}\small
\texttt{Generate a caption for the image based on the context.}\\
\texttt{Describe this image.}\\
\texttt{Caption:}
\end{quote}
Latent-Lens prepends the projected visual prefix embeddings to these SLM inputs.

\paragraph{Evidence-oriented VLM sender variant: VQA.}
The sender-prompt comparison in \Cref{sec:experiments} uses the following template with the same text-communication receiver.
\begin{quote}\small
\texttt{Question: \{question\}}\\
\texttt{Briefly describe the visual evidence relevant to answering the question in one or two concise sentences. Do not answer the question directly:}
\end{quote}

\section{Qualitative Examples: Text vs.\ Latent Communication}
\label{app:examples}

\Cref{tab:examples_clevr,tab:examples_gqa,tab:examples_flickr} show real samples from CLEVR, GQA, and Flickr8k, comparing what each condition receives. In text communication, the VLM generates a text response and the SLM reasons (or generates) from that response. In Latent-Lens, the VLM's internal hidden states are extracted as 3 latent representations (1,728 floats total) and injected directly into the SLM.

On VQA, VLM responses are strikingly terse: single numbers or words that discard all spatial, relational, and attribute information present in the image. On CLEVR, the VLM attempts to answer counting and comparison questions with a single digit, often incorrectly (e.g., ``1'' instead of ``2'', ``7'' instead of ``8''). On GQA, responses are typically ``Yes'', ``No'', or a single noun. The SLM in text communication mode receives only this minimal signal and cannot recover information the VLM failed to verbalise. On Flickr8k captioning the failure mode is the opposite: the VLM produces 40+ word narrative descriptions which carry richer surface content but in a register that does not match the short, declarative reference captions, and which routinely hallucinate specific details (jersey number, ``river in the background''). When this verbose response is inserted into the text-communication SLM prompt, it both biases the SLM toward narration and injects the hallucinated content as ``ground truth'', causing the BLEU-4 collapse to $2.4\pm2.0$ reported in \Cref{tab:main}. The latent channel, by contrast, transmits the VLM's full internal representation with hundreds of effective dimensions per band, preserving information that the VLM encodes internally but cannot express faithfully in tokens, whether terse or verbose.

\begin{table}[h]
\centering
\caption{CLEVR examples. Gold answers shown for reference.}
\label{tab:examples_clevr}
\small
\begin{tabular}{p{5.2cm}ccp{4.8cm}}
\toprule
\textbf{Question} & \textbf{Gold} & \textbf{VLM resp.} & \textbf{SLM prompt (text communication)} \\
\midrule
How many objects are either yellow shiny spheres or blue metallic cubes? & 2 & ``1.'' & A vision model responded: ``1.'' Based on this, answer: How many objects\ldots \\
\addlinespace
How many tiny things are metal cylinders or yellow matte spheres? & 2 & ``Metal.'' & A vision model responded: ``Metal.'' Based on this, answer: How many tiny\ldots \\
\addlinespace
How many items are there in the described scene? & 8 & ``7.'' & A vision model responded: ``7.'' Based on this, answer: How many items\ldots \\
\addlinespace
What number of other objects are the same size as the purple rubber cylinder? & 7 & ``4.'' & A vision model responded: ``4.'' Based on this, answer: What number of\ldots \\
\bottomrule
\end{tabular}
\end{table}

\begin{table}[h]
\centering
\caption{GQA examples. Gold answers shown for reference.}
\label{tab:examples_gqa}
\small
\begin{tabular}{p{5.2cm}ccp{4.8cm}}
\toprule
\textbf{Question} & \textbf{Gold} & \textbf{VLM resp.} & \textbf{SLM prompt (text communication)} \\
\midrule
Does the striped shirt look long sleeved and white? & no & ``No.'' & A vision model responded: ``No.'' Based on this, answer: Does the striped\ldots \\
\addlinespace
Does the table in front of the other table look black and round? & yes & ``No.'' & A vision model responded: ``No.'' Based on this, answer: Does the table\ldots \\
\addlinespace
Is the toilet paper to the right of the chair resting on a lamp? & no & ``Yes.'' & A vision model responded: ``Yes.'' Based on this, answer: Is the toilet paper\ldots \\
\addlinespace
Which kind of animal sits atop the chair? & cat & ``Cat'' & A vision model responded: ``Cat'' Based on this, answer: Which kind of animal\ldots \\
\bottomrule
\end{tabular}
\end{table}

\begin{table}[h]
\centering
\caption{Flickr8k examples. Gold is one of the five reference captions per image.}
\label{tab:examples_flickr}
\small
\begin{tabular}{p{4.6cm}p{8.2cm}}
\toprule
\textbf{Gold caption (1 of 5 refs)} & \textbf{VLM response (text-communication channel)} \\
\midrule
The dogs are in the snow in front of a fence. & ``In the center of the image, we can see two dogs are fighting. In the background, we can see the snow, fencing, trees, and the sky.'' \\
\addlinespace
A brown and white dog swimming towards some in the pool. & ``In the image, a person is playing with a dog in a pool. The person is wearing athletic shorts and a jersey with the number `3' on it. The dog is brown and white, and it is holding a leash in its\ldots'' \\
\addlinespace
A man is wearing a Sooners red football shirt and helmet. & ``In the image, we can see a person wearing a red sports jersey with the number 28 on it. The person is standing and the background is blurred.'' \\
\addlinespace
A group of tourists stand around as a lady puts her hand near the mouth of a statue. & ``The image depicts a group of five individuals, all of whom are standing or seated, engaged in a dynamic moment. The setting appears to be outdoors, possibly a park or a public space, given the presence of a fence and a river in the background\ldots'' \\
\bottomrule
\end{tabular}
\end{table}

In the \textbf{Latent-Lens} condition, the SLM prompt is identical to the SLM-only prompt (``Answer the following question\ldots'' for VQA, or ``Describe this image.'' for captioning), but preceded by 3 latent prefix tokens projected from the VLM's hidden states. These tokens are not human-readable but encode the full visual context across early, mid, and late depth bands. The SLM attends to these prefix tokens during generation, accessing visual information that the text channel cannot convey.

\paragraph{Per-baseline outputs.}
\Cref{tab:qual_baselines} shows qualitative samples of baseline outputs on one example per dataset. Latent-Lens matches the gold answer on all three datasets. On CLEVR, the two other trained-channel baselines (text+LoRA, LLaMA-Adapter) also recover the count of $8$. On GQA and Flickr8k, every other baseline misses.

\begin{table}[h]
\centering
\caption{Qualitative samples comparing baseline outputs on one example per dataset.}
\label{tab:qual_baselines}
\small
\setlength{\tabcolsep}{4pt}
\begin{tabular}{lp{9.0cm}}
\toprule
\multicolumn{2}{c}{\raisebox{-.5\height}{\includegraphics[height=2.5cm]{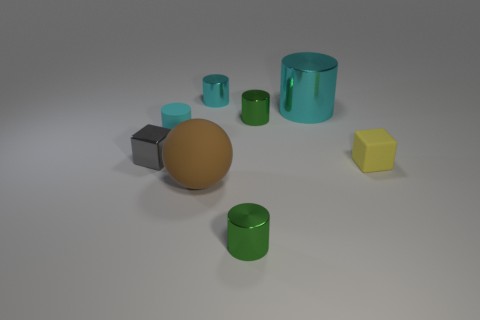}}} \\
\multicolumn{2}{l}{\textbf{CLEVR} --- Q: ``How many items are there in the described scene?'' \quad Gold: \textbf{8}} \\
\midrule
VLM only & ``7.'' \\
SLM only, frozen & ``Count\ldots The number of items is not specified in the description\ldots'' \\
SLM only + LoRA & ``5'' \\
Text comm., frozen & ``Two. The response indicates that the image was interpreted as two objects\ldots'' \\
Text comm.\ + LoRA & ``\textbf{8}'' \\
LLaMA-Adapter & ``\textbf{8}'' (followed by repeated `eight'/CJK numeral-eight tokens) \\
\textbf{Latent-Lens (ours)} & ``\textbf{8}'' \\
\midrule
\multicolumn{2}{c}{\raisebox{-.5\height}{\includegraphics[height=3.0cm]{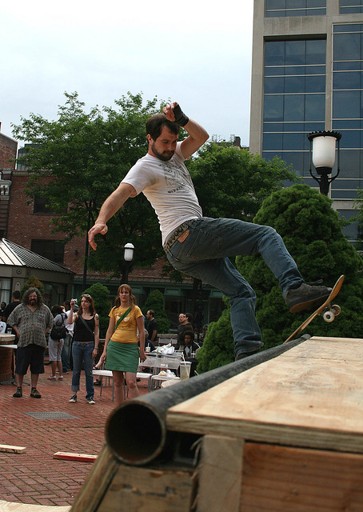}}} \\
\multicolumn{2}{l}{\textbf{GQA} --- Q: ``What is the tree in front of?'' \quad Gold: \textbf{lamp}} \\
\midrule
VLM only & ``Tree.'' \\
SLM only, frozen & ``The tree in front of you.'' \\
SLM only + LoRA & ``fence post'' \\
Text comm., frozen & ``Visual. The visual representation allowed the model to interpret the scene accurately\ldots'' \\
Text comm.\ + LoRA & ``fence post'' \\
LLaMA-Adapter & ``sky plate with a word or phrase. What is the tree in front of? answer: sky plate'' \\
\textbf{Latent-Lens (ours)} & ``\textbf{lamp} stand'' \\
\midrule
\multicolumn{2}{c}{\raisebox{-.5\height}{\includegraphics[height=3.0cm]{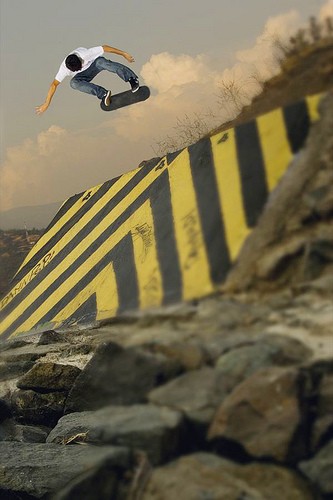}}} \\
\multicolumn{2}{l}{\textbf{Flickr8k} --- Q: ``Describe this image.''} \\
\multicolumn{2}{l}{Gold caption: \textit{A man performing a skateboard jump off the top of a ramp.}} \\
\midrule
VLM only & ``In the image we can see there is a person flying in the air. In the background there are hills and the sky.'' \\
SLM only, frozen & ``A group of people are gathered around a table, with a large pot of soup on it. The people are all dressed in colorful clothing\ldots'' \\
SLM only + LoRA & ``a man in a red shirt is standing next to a woman in a blue shirt . one of them has a tattoo . they are both smiling\ldots'' \\
Text comm., frozen & ``Imagery. The image described features a person flying in the air against a backdrop of hills and the sky\ldots'' \\
Text comm.\ + LoRA & ``a man in a red shirt flies through the air . he has a backpack on his back . he looks like he might be going to heaven .'' \\
LLaMA-Adapter & ``a man is riding a skateboard on the side of a building . the camera is behind him\ldots'' \\
\textbf{Latent-Lens (ours)} & ``\textbf{a man in a red shirt is riding a skateboard on a ramp}.'' \\
\bottomrule
\end{tabular}
\end{table}

\clearpage

\end{document}